*Article*

# Vibration Fault Diagnosis in Wind Turbines based on Automated Feature Learning

Angela Meyer[1],[*]

1 Bern University of Applied Sciences, Bern, Switzerland
* Correspondence: angela.meyer@bfh.ch

**Abstract:** A growing number of wind turbines are equipped with vibration measurement systems to enable a close monitoring and early detection of developing fault conditions. The vibration measurements are analyzed to continuously assess the component health and prevent failures that can result in downtimes. This study focuses on gearbox monitoring but is applicable also to other subsystems. The current state-of-the-art gearbox fault diagnosis algorithms rely on statistical or machine learning methods based on fault signatures that have been defined by human analysts. This has multiple disadvantages. Defining the fault signatures by human analysts is a time-intensive process that requires highly detailed knowledge of the gearbox composition. This effort needs to be repeated for every new turbine, so it does not scale well with the increasing number of monitored turbines, especially in fast growing portfolios. Moreover, fault signatures defined by human analysts can result in biased and imprecise decision boundaries that lead to imprecise and uncertain fault diagnosis decisions. We present a novel accurate fault diagnosis method for vibration-monitored wind turbine components that overcomes these disadvantages. Our approach combines autonomous data-driven learning of fault signatures and health state classification based on convolutional neural networks and isolation forests. We demonstrate its performance with vibration measurements from two wind turbine gearboxes. Unlike the state-of-the-art methods, our approach does not require gearbox-type-specific diagnosis expertise and is not restricted to predefined frequencies or spectral ranges but can monitor the full spectrum at once.

## 1. Introduction

The globally installed wind power capacity is constantly growing thanks to international efforts to limit the global mean temperature rise by replacing fossil fuels [1]. A major fraction of the levelized cost of wind energy consists of the operation and maintenance costs of the wind farms [2]. Continuous health monitoring of wind turbine components forms an important part of the work of wind farm operators as it helps to limit the extent of unforeseen maintenance costs. To reduce the operation and maintenance costs of their wind farms, many operators and asset managers are applying remote condition monitoring techniques to detect incipient faults before they can result in major damage.

Gearboxes are among the most critical and costly components to replace in a wind turbine in terms of the equipment, replacement work and downtime costs per failure [3-6]. Therefore, a growing number of wind turbine gearboxes is being equipped with vibration measurement systems to enable a close monitoring and early detection of developing fault conditions in the gearbox components [7-9]. The vibration monitoring signals require analysis and interpretation to prevent failures. Numerous approaches have been proposed to assess the vibration signals from wind turbine gearboxes in the time and frequency domains. Examples include the time-domain monitoring of waveform features such as root mean square deviations, peak-to-peak amplitudes, and kurtosis. In the frequency domain, methods such as spectral line analysis, envelope and sideband analysis have been proposed [8-11]. Thus, the state-of-the-art vibration diagnostics methods applied for wind turbine (WT) gearboxes in practice rely on the extraction of hand-crafted features from the gearbox vibration signals. The features need to be defined by a human analyst before they can be extracted. Only after they have been defined and extracted from the vibration measurements, the features may be used to infer information about potential faults in gearbox components based on statistical methods or machine learning models [12-13]. Typical handcrafted features that are in use for WT gearbox fault diagnostics are the position and amplitude of spectral lines corresponding to characteristic frequencies of gearbox components, such as gear mesh frequencies, and other characteristic metrics such as the root mean square deviation and kurtosis of parts of the vibration time series [8-11]. However, these state-of-the-art vibration diagnostics methods have multiple disadvantages. First, they require a labor-intensive upfront conception and handcrafting of feature definitions, which constitutes a significant time and workforce effort. Second, many state-of-the-art approaches and feature definitions need a highly detailed knowledge of the gearbox type, manufacturer, composition and dimensions, its bearing and gear types, gear teeth numbers, and so on. This information needs to be gathered for every single gearbox before the start of the monitoring. As a result, the state-of-the-art fault diagnostics and feature extraction approaches can generally not be transferred straightforwardly to new turbine types added to an operator's wind power portfolio. For every new turbine entering the portfolio, detailed turbine composition information needs to be collected from the manufacturer and the vibration features need to be reviewed, adapted and extracted. This constitutes a large resource-intensive initial effort that many wind farm operators and asset managers are hesistant to make. Third, after a feature definition and extraction method has been implemented, thousands of characteristic spectral values per turbine gearbox need to be stored and monitored, which requires costly storage resources and computing time in the remote monitoring centers of the turbine operators and asset managers. Fourth, the state-of-the-art approaches do not analyse the full vibration spectrum but focus on monitoring only isolated aspects thereof, such as a set of characteristic frequencies, or they focus on global metrics of the vibration time series or spectrum. Unlike the proposed approach, they do not support an automated simultaneous vibration monitoring of the full spectral range. Lastly, features defined by human analysts can lead to imprecise decision boundaries and less accurate fault diagnostics predictions than features that have been learnt by the machine learning algorithms themselves [14]. The state-of-the-art feature definition and extraction methods may result in lower diagnostics accuracy, more false alarms and false negatives, especially in ambiguous boundary cases that require additional inspection and decision making by remote monitoring staff, than fault diagnostics methods which learn and extact the optimal features themselves. A reliable feature definition and extraction is essential to the fault diagnostics process. For an illustrative example of how the chosen upfront feature definitions can affect the fault detection quality, we refer to the study presented by [11].

The research gap addressed by this study is the development of a fault diagnostics method for vibration-monitored wind turbine gearboxes that

1) learns and extracts an optimal set of discriminative features in an automated manner, not requiring any feature engineering,

2) analyses the full vibration spectrum, rather than focusing only on isolated pre-defined aspects thereof, and

3) that is even applicable if only few fault observations are available.

Consequently, the objective of this paper is to introduce and demonstrate a novel fault diagnostics approach for vibration-monitored wind turbine gearboxes that can overcome the discussed disadvantages of the state-of-the-art methods. In particular, the novel approach is expected to learn optimal discriminative features in an automated manner and classify the gearboxes' health conditions based on these features without requiring any human feature definition and extraction. It is also expected to analyse the full vibration spectrum and be applicable even in situations where sufficient model training data for fault-type classification are unavailable.

This paper is organized as follows. Section 2 introduces the proposed fault diagnostics approach. Section 3 describes the method applied and data employed in a gearbox failure case study, whereas section 4 discusses the analysis and results. Our conclusions are presented in section 5.

## 2. Fault Diagnosis Method

The proposed fault diagnosis method comprises two stages. The first stage performs unsupervised anomaly detection on the features learnt and extracted from spectrograms of each monitored gearbox component (Figure 1). This stage one accounts for the fact that many WT operators have access to only few or even no sensor measurements from actual gearbox fault incidents as these are relatively rare events and they can arise from a range of different causes. While methods that require labelled observations of gearbox faults (as in proposed stage two below) may be less beneficial to such operators, anomaly detection methods based on measurements taken in normal healthy operation state will still be available and highly useful to them even in absence of labelled fault observations.

The second stage of the presented approach employs a multi-label classification method to diagnose specific gearbox fault types based on past fault observations (Figure 2). This stage mainly benefits operators who have access to measurements from observed gearbox faults that enable training a corresponding fault type classification model. Therefore, the proposed stage two is performed only if sufficient fault observations are available to the operator's remote monitoring staff in charge of implementing the proposed approach.

The fault diagnoses are made based on features extracted from vibration spectrograms. To this end, vibration measurements are taken continuously from numerous accelerometer-monitored gearbox components and are accessed through the turbine's condition monitoring system (CMS). The accelerometer measurement time series are subjected to short-time Fourier transforms (STFT) to monitor the temporal evolution of the vibration spectra in the time-frequency domain. The resulting spectrograms from all accelerometer-monitored gearbox components serve as inputs to feature extraction neural networks composed of convolutional and pooling layers as described below. Unlike the state-of-the-art fault diagnostics methods, the proposed approach does not require gearbox-type specific information. Therefore, it can be introduced to even large WT portfolios without the upfront efforts and investments required for existing methods.

The operators will be informed both in stages one and two in case a significant deviation from healthy component spectrograms has been diagnosed. Importantly, both stages of the proposed fault diagnostics approach rely on automated feature learning and extraction that is performed by an algorithm rather than a human analyst. This is achieved by the application of convolutional neural networks (CNNs) [14-15], as shown in Figures 1 and 2. We refer to [16] for a technical introduction to CNNs. CNNs were selected for the proposed health state classification approach because, unlike other models, they are capable of learning and extracting features from image data without human assistance.

CNNs are computational models that are capable of extracting the relevant features without any human assistance. They accomplish this by learning optimal convolutional filters based on historical training data, in this case vibration measurements and fault observations. Thanks to this property, CNNs have enabled major performance improvements in fields such as speech recognition and object detection in recent years [14-15]. CNNs are artificial neural networks that consist of convolutional and pooling layers trained to perform feature learning and extraction based on past observations. These layers are subsequently linked to fully connected layers to perform desired classification or regression tasks based on the previously extracted features. During model training, the CNN weights optimization algorithm effectuates automated feature learning and extraction to construct a low dimensional representation of the input spectra which is subsequently fed to the anomaly detection model (stage one, Figure 1) or the fault type classification network (stage two, Figure 2).

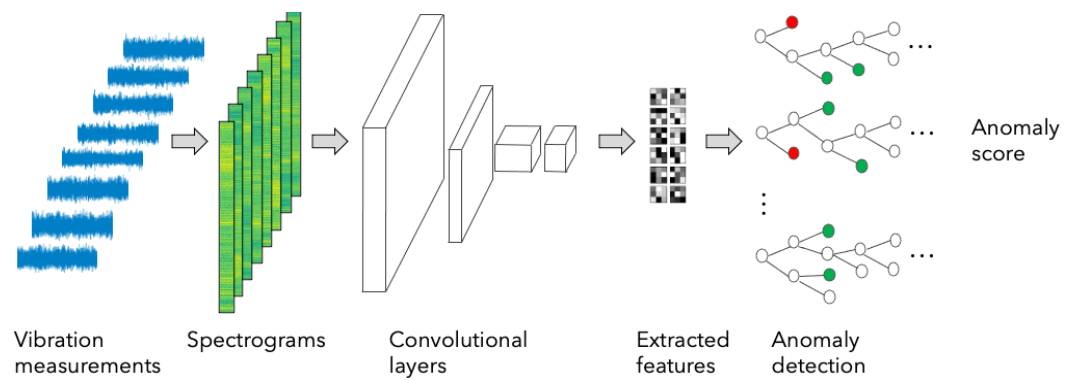

**Figure 1.** Stage one of the proposed method. Anomaly detection based on features extracted by convolutional and pooling layers from spectrograms of the monitored components in their healthy state. An isolation forest model is trained on the extracted features. The trained model is subsequently employed for detecting anomalous spectra, indicated as red nodes in the isolation trees shown on the right-hand side of the chart.

In the first stage of the presented fault diagnostics approach, the feature extraction part is succeeded by an isolation forest model (Figure 1) for detecting anomalous spectrograms in an automated manner. The extracted features serve as input to the isolation forest algorithm [17] that is adapted to distinguish anomalous from normal spectrograms with regard to the component health state based on historical accelerometer measurements. The isolation forest algorithm identifies potential anomalies by how quickly the input spectrograms can be isolated from the rest of the spectrograms using a decision-tree-based approach. A health-state classifier is trained using examples from only one class, namely observations from healthy gearbox components only. This is a highly relevant scenario in practice because fault observations of WT gearbox components are often lacking: Wind farm operators usually have a large amount of CMS sensor observations from different parts of the drive train from multiple months or years of operation. Typically, the vast majority of these measurements from the CMS system are taken under normal operating conditions from healthy components. Fault conditions and damages occur relatively rarely in commercial turbines that are operated and maintained in accordance with the manufacturer's recommendations. Therefore, there is a relative lack of such fault observations. This lack strongly restricts the training and application of machine learning models for fault type classification because those models require a significant amount of training data. Therefore, machine learning models trained only on observations from healthy gearbox components tend to be more widely applicable in practical applications and are highly relevant when comprehensive fault observations are lacking.

To train the health-state classifier using only observations from healthy gearbox components, we compared two anomaly detection approaches: the isolation forest algorithm introduced above and one-class support vector machines [18]. The isolation forest approach is known for its fast computational training time [17]. In the case study presented below, it outperformed the one-class SVM by more than a factor of 30 in terms of required training time but provided no advantages in terms of prediction performance. Therefore, our discussion of the case study below focuses on the developed isolation forest model with its more attractive training times and accordingly larger practical relevance.

In stage two of the proposed fault diagnostics method, our goal is to train a health-state classifier for diagnosing specific fault conditions in gearbox components using both accelerometer measurements from healthy components and from damaged ones. The features extracted by the convolutional and pooling layers serve as input to train multi-label fault type classifiers, as illustrated in Figure 2.

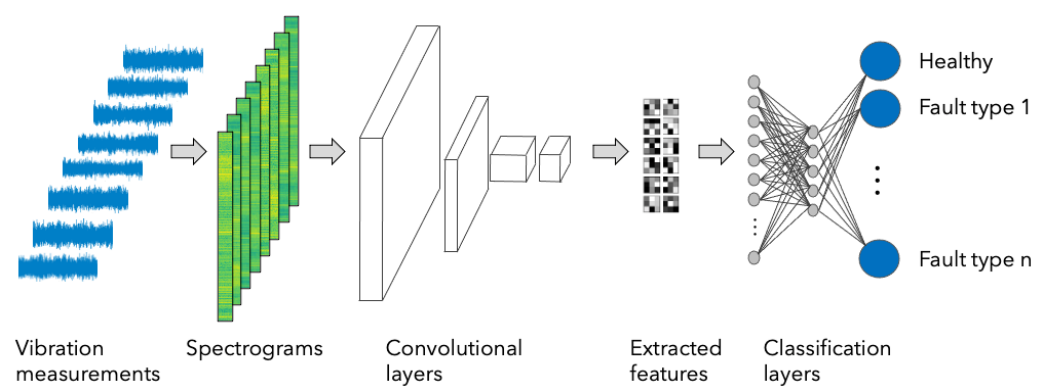

**Figure 2.** Stage two of the proposed approach. Again, the feature learning and extraction is performed autonomously by convolutional and pooling layers. The proposed fault diagnostics model for the health state classification in wind turbine gearboxes is a fully connected multi-label neural network, as shown on the right-hand side of the chart.

Multi-label classification [19-21] is especially beneficial when accelerometer measurements taken during evolving and evolved past gearbox damage are available for the gearbox types of interest. A multi-label classification model is estimated for monitoring multiple gearbox components simultaneously based on high-frequency acceleration measurements from accelerometers attached in close proximity of the monitored components. The multi-label classification enables a joint classification of multiple damage types, wherein each data instance is simultaneously assigned multiple labels. Each label indicates the membership status in one of multiple classes in a binary manner. More formally, a multi-label classification algorithm estimates a map $f: \mathbb{R}^k \to \mathbb{R}^n$ based on a training set $\{(x_{1,i}, \ldots, x_{k,i}, y_{1,i}, \ldots, y_{n,i}), i = 1, \ldots, m\}$ of size $m$ wherein the coordinates of any $y \in \mathbb{R}^n$ can take binary values only, $y_j \in \{0, 1\} \, \forall \, j = 1, \ldots, n$.

## 3. Case study

The proposed fault diagnosis approach is demonstrated and tested based on vibration measurements from multiple gearbox components taken while the gearbox was operating on a test rig. The proposed approach is demonstrated and its performance tested using accelerometer measurements from the gearboxes of two 750 kW wind turbines operated on a WT test rig at the National Renewable Energy Laboratory (NREL). The data were collected by NREL for its Gearbox Condition Monitoring Round Robin study [23]. The accelerometer measurements were taken over a ten-minute period from two identical gearboxes of the 750kW wind turbines with one of the gearboxes in a healthy unimpaired state and the other gearbox suffering from multiple damaged components after an oil loss event which had caused moderate damage to the gearbox gears and bearings. Each of the two gearboxes has a transmission ratio of 1:81.491 and comprises a low-speed planetary stage and two parallel stages. Figure 3 shows a schematic of the gearbox. We refer to [23-24] for a detailed description and visualization of the test stand, gearbox, monitoring system and measurement setup.

To demonstrate the fault diagnostics approach and automated feature learning and extraction, the following components were selected for this case study in an arbitrary manner from among the set of components monitored in the NREL study [23]: Accelerometer 1 was attached to the ring gear (component 1) at a bottom-facing location in both the healthy and the damaged gearbox to measure radial accelerations. Accelerometer 2 was attached to the low-speed shaft bearing (component 2) also to measure radial accelerations, whereas accelerometer 3 monitored the radial accelerations of the high-speed shaft downwind bearing (component 3). All three components - the ring gear, the low-speed shaft bearing and the high-speed shaft bearing - exhibited different degrees of damage in the damaged gearbox such as scuffing. In total, readings from six sensors are considered. The accelerometer measurements at the undamaged gearbox and at the damaged gearbox were each taken at 40 kHz sampling frequency under constant load and speed of 22.09 rpm low-speed shaft and 1800 rpm high-speed shaft speed for a duration of 10 minutes.

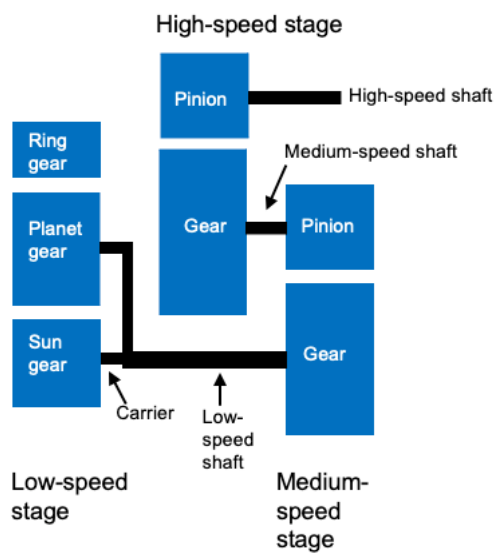

**Figure 3.** Schematic of the gearbox (based on Sheng, 2012)

## 4. Discussion

To create the spectrograms that will be input to the CNN for feature extraction, the accelerometer measurements were split into four segments per second and a separate short-time Fourier transform (STFT) was computed for each segment. A short-time Fourier transform enables the frequency analysis of a signal as it changes over time [25-26]. The length of four segments per second was selected by investigating the tradeoff between time and frequency resolution so as to maintain a high frequency resolution and sufficient temporal resolution, as shown in Figure 4. The frequency resolution should be sufficient for resolving typical spectral differences arising between healthy and damaged states of the monitored components. The STFTs were computed with an overlap of 0.2 seconds for adjacent segments. However, the length of the overlap had no significant effect on the performance of the subsequent anomaly detection and classification models. Given the 40 kHz sampling rate, vibration frequencies of up to the Nyquist frequency of 20 kHz can be resolved. For the present fault diagnostics case study, we focused the analysis on vibration frequencies of up to 1 kHz.

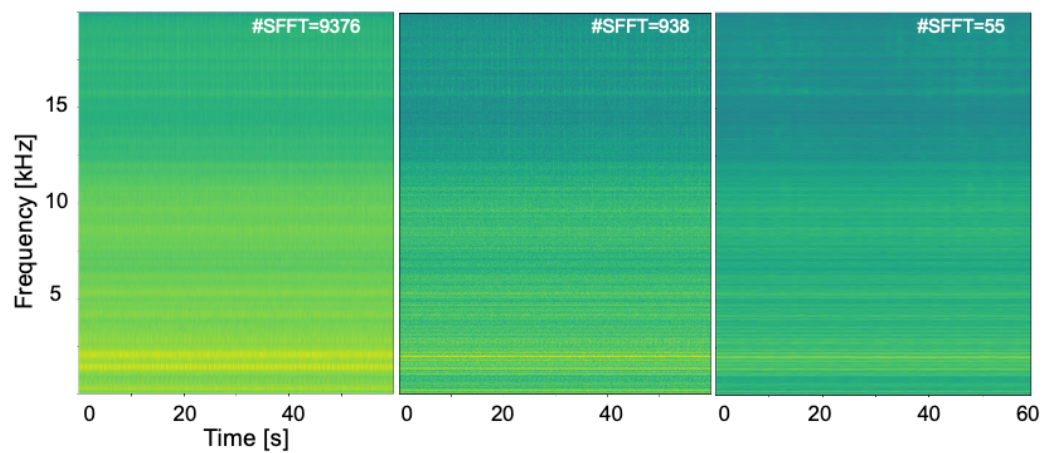

**Figure 4.** Tradeoff between frequency and temporal resolution depending on the number of short-time Fourier transforms performed per minute (#SFFT) for creating the spectrograms. The longer the input time series to the Fourier transform, the higher the frequency resolution achieved.

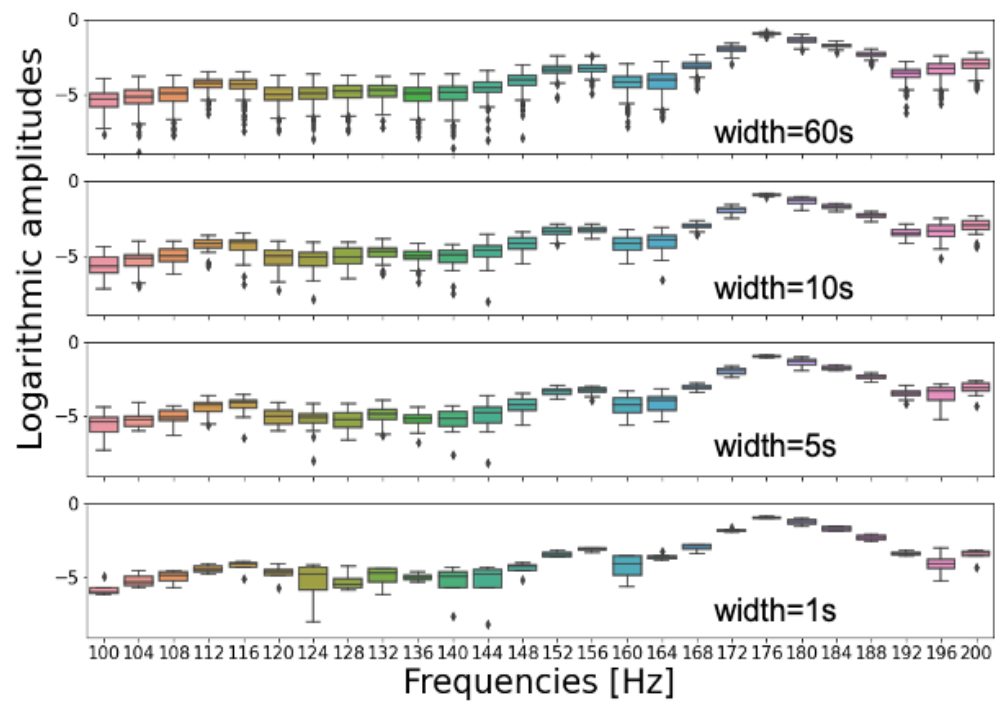

**Figure 5.** One-second intervals were found to be sufficiently representative of the STFT amplitude variability when examining this variability over a range of intervals, including 1, 5, 10 and 60 seconds as shown in the subpanels, for all frequencies up to 20 kHz. For illustration, the amplitude variability is only shown for frequencies in the range of 100 to 200 Hz.

Prior to the model training, we sampled segments from the resulting spectrograms with replacement in order to augment the training dataset. We sampled segments of one second in length to ensure that short vibration measurement periods (of only one second) will be sufficient as input to the fault diagnostics model when it is used for inference in a condition monitoring software or CMS system. One-second intervals were found to be sufficiently representative of this amplitude variability when examining the STFT amplitude variability over time for all frequencies up to 20 kHz, as shown in Figure 5. To test the sensitivity of the presented fault diagnostics approach with regard to the temporal length of the sampled spectrogram segments, we performed our analysis for spectrograms with time lengths ranging up to 6 seconds, finding that this choice did not significantly affect the results. After sampling the one-second segments from the vibration spectra of healthy and faulty components, the resulting dataset was randomly shuffled and partitioned into training, validation and test sets. The training set in this case study contained 80000 instances. The classification method was validated using a validation set of vibration measurements from healthy and damaged components based on 10000 instances. The test set also contained 10000 instances.

Figure 6 shows subsets of the spectrograms derived from the vibration measurements of the three accelerometers at the healthy and the damaged gearbox components. Two healthy and two damaged instances are shown for each of the three components for illustration. As can be seen in the figure, the spectral differences between vibration measurements from the healthy and the damaged components have been sufficiently resolved by the Fourier transforms to enable discriminative feature learning.

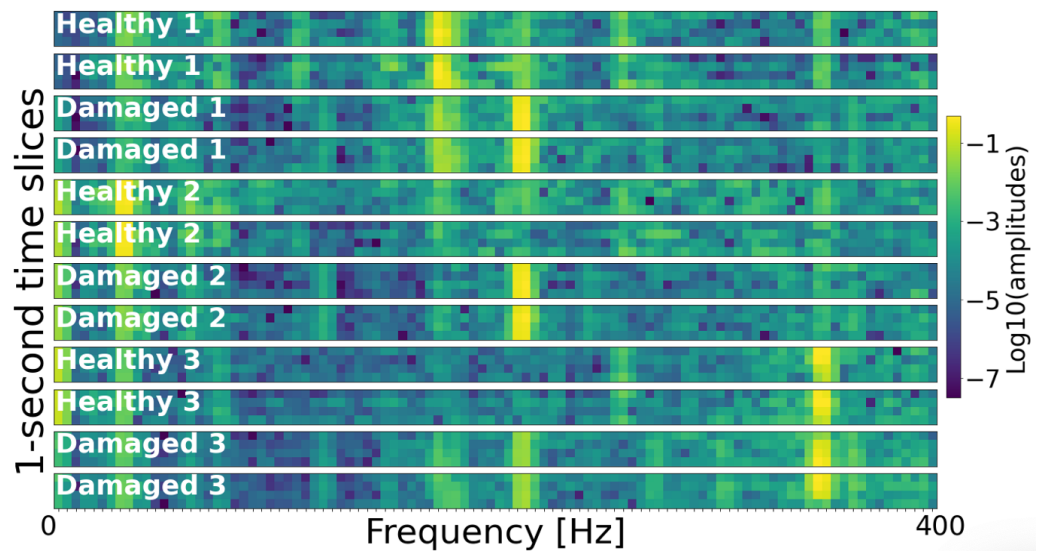

**Figure 6.** Subsets of the spectrograms derived from the vibration measurements of the three accelerometers at the healthy and the damaged gearbox components. Two healthy and two damaged instances are shown for each of the three components. For instance, "Healthy 1" labels two different 1-sec spectrogram segments of component 1 in healthy state. Each spectrogram segment runs over 1 second and displays logarithmic STFT amplitudes for frequencies up to 400 Hz.

**Stage 1: Isolation forests for detecting anomalous vibration spectra.** The spectrograms prepared as outlined above served as input to convolutional and pooling layers of a CNN that learned and extracted discriminative features in an automated manner. For the feature learning and extraction, we defined a network architecture consisting of one convolutional layer with 16 convolutional filters of 3-by-3 pixels followed by a max pooling layer with a 2-by-2 pixels window size and batch normalization [27]. This architecture is of low complexity and also enabled a high classification accuracy in the multi-label classification of stage two of the presented fault diagnostics approach.

The extracted features were input to an isolation forest algorithm for one-class classification [17] which identifies spectral anomalies based on how hard it is to isolate a particular spectrum from the rest of the spectra in the training set. A forest containing 100 isolation trees was trained on the extracted features. The model parameters are summarized in Table 1. The training dataset contained features from the spectrograms of only healthy components. We specified the fraction of anomalous data instances estimated to be present in the training data to less than one over the training set size. The anomaly score computed by the model for each training, validation and test set instance corresponds to the number of splits averaged across the isolation tree forest that is needed to isolate a data point (Figure 1). Thus, the anomaly score is the average path length from root to leaf node in an isolation tree. As shown in Figure 7, the spectra from the healthy and damaged components are clearly separable using the computed anomaly scores, so the isolation forest model is well suited for identifying components with anomalous health states even for high dimensional feature spaces as in the present case study.

The health state discrimination is performed in an unsupervised manner to make it applicable to WT operators whose remote condition monitoring team does not have sufficient amounts of fault observations. Since we are actually in possession of such observations in this case study, we employed a test dataset that had not been used in the model training (Figure 7) and estimated performance metrics based on the test dataset. We found both recall and precision to be 100% on the test dataset for the proposed isolation forest approach and model architecture. Recall is a performance metrics that designates the fraction of true positives over all actual positives, in this case the fraction of all correctly identified instances of a given fault type over all actual occurrences of that fault type. Precision is an alternative performance metrics that denotes the fraction of true positives over all instances there were identified as positive. In other words, the precision states what fraction of the identified observations of a given fault type were correctly identified as observations of that fault type.

**Table 1.** Parameters of the trained isolation forest model (stage one) and of the convolutional neural network (stage two).

| Model | Model parameter |
| --- | --- |
| Isolation forest | Number of isolation trees equals 100; Threshold for outliers fraction equals 0.0001 |
| Convolutional neural network | One convolutional layer with four 3-by-3 filters followed by a 2-by-2 max pooling layer and batch normalization, followed by a dense layer of four fully connected nodes and a 3-node output layer, 10% drop out rate |

We repeated the analysis with the same feature learning and extraction architecture employed in stage two below, which naturally resulted in the same extracted feature set as used in the multi-label classification step. Specifically, this architecture comprised one convolutional layer with only four convolutional 3-by-3 filters followed by a 2-by-2 max pooling layer and batch normalization. As before, the extracted features were then input to the isolation forest algorithm for detecting anomalous spectra. The change in feature extraction architecture had no significant effect on the spectra's anomaly scores (Figure 67) and also resulted in 100% recall and precision.

In addition, one-class support vector machines (SVMs) [18] were investigated as a further approach for the vibration-based anomaly detection in this study. However, the SVM algorithm required more than 30 times more model training time than the isolation forest on the same training set and processor, an AMD EPYC 7B12 2.25 GHz processing unit, though no improvement of detection performance was observed.

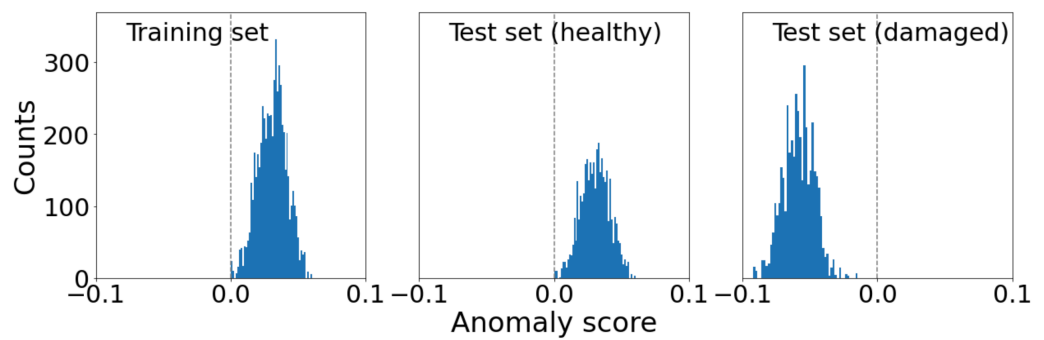

**Figure 67.** Anomaly detection based on anomaly scores computed with the isolation forest algorithm. The training set contains only spectra from healthy component states, corresponding to absence of anomalies as indicated by a positive anomaly score (left panel). We tested the model using additional spectra from healthy component states (middle panel) and from damaged component states (right panel). In the latter case, the anomaly scores were negative, indicating anomalies as expected. Healthy and damaged components are clearly separable with this approach.

**Stage 2: Multi-label classification for fault-type diagnostics.** As in stage one, the spectrograms were subjected to convolutional and pooling layers to enable feature learning and extraction based on the training dataset. Subsequently, a fault-type classification was performed with the extracted features using fully-connected neural network layers. Jointly, the convolutional, pooling and fully connected layers established the convolutional neural network for the multi-label fault type classification. Once trained, the CNN predicts the probabilities of all three fault types considered in this case study based on a given spectrogram to be diagnosed.

To arrive at the final CNN structure (Table 1), we started from a more complex CNN architecture and successively reduced the number of convolutional and pooling layers, filters and fully connected layers while maintaining maximal validation set accuracy. We selected the least complex CNN architecture that could achieve the highest possible classification accuracy on the validation set. In this case study, the resulting CNN architecture comprises one convolutional layer with four 3-by-3 filters followed by a 2-by-2 max pooling layer and batch normalization. This first part ensures the learning and extraction of features based on which the subsequent classification can be performed. Two fully connected layers were added to the network and a 10% dropout rate applied to avoid overfitting the training data [28]. The first fully connected layer comprised four nodes and the output layer consisted of three nodes with a sigmoid activation function for the output layer. The model predicts three binary labels, one of which for each of the monitored gearbox components, and indicates whether or not a fault has been detected in the respective component. The output layer with the three neurons and the sigmoid activation functions is providing probabilities for a given spectrum to belong to a particular fault-type class. The model parameters were determined iteratively with the Adaptive Moment Estimation (Adam) optimization algorithm [29] by optimizing a binary cross-entropy loss function. In doing so, multiple binary classification decisions can be optimized at once. The model training was performed for 20 epochs and batch size equal to 32. Different batch sizes did not affect the classification accuracy. Logarithmic transformations of the input spectra had no effect either on the classification accuracy. This architecture enabled a 100% classification accuracy for all three component fault type classes on the validation and test sets. We arrived at this architecture through a grid search by starting from a more complex CNN with the number of nodes equal to powers of two, and then reducing the network complexity while maintaining the high accuracy of 100% on the validation set, as described above. The test set classification accuracy of 100% was achieved both with and without the logarithmic transformation of the spectrogram segments inputs to the convolutional and pooling layers (Figure 6). The models trained as part of the hyperparameter optimization converged to a loss function minimum within 20 epochs without overfitting.

With regard to the limitations and the future research needs arising from the present study, we point out that, first, all acceleration measurements in the present study were taken under constant speed and load conditions on a test stand. The introduced fault detection approach should be field tested under variable speed and load conditions in future work. In practice, the wind speed driving the turbine is fluctuating which results in a variable load and shaft speed and may cause frequency smearing in the spectral representation [9]. However, this condition can be overcome by synchronizing the measurements with the wind turbine's rotational speed, for instance by sampling under identical wind and load conditions. Second, for applications in operating wind turbines, the performance of the fault type classification model should also be investigated for a significantly larger number of monitored components and fault types and for damage processes that are evolving over time. This investigation will require more comprehensive field or laboratory measurement datasets. Third, attention also needs to be paid to avoiding possible data imbalance issues when training a fault diagnostics model. While this does not affect stage one of the proposed fault diagnosis approach, it may be relevant in the application of the methods introduced for stage two. Data imbalance refers to situations where there are disproportionate numbers of observations in the output classes. For instance, there may be a large number of observations for one class, say fault type 1, but only few observations for another class. In the presented case study, all fault types were represented with similar numbers of observations. This may not always be the case in practice. Typically, more vibration measurements will be available from healthy components because fault situations are less common than WT gearboxes operating in normal health state. Vibration measurements from damaged components or components in which a damage starts to develop are typically in the underrepresented class. One method to address data imbalance is by over- or undersampling to arrive at an augmented and more balanced training dataset. This may be achieved, for example, by random resampling with replacement (statistical bootstrapping) from the available fault observations, so that all monitored WT components are equally represented, both in healthy and damage states, in the training, validation and test datasets. This approach relies on the assumption that the data used for the bootstrapping are sufficiently representative of the underlying data-generating process. A more comprehensive discussion of methods for addressing data imbalance is not in the scope of this work and we refer to the work of other authors, e.g. [30].

## 5. Conclusions

An increasing number of wind turbines are equipped with vibration-measurement systems to enable a close monitoring and early detection of developing fault conditions in the gearbox. Gearboxes are among the most critical and costly components to replace in wind turbines. The current state-of-the-art gearbox fault diagnostics algorithms rely on upfront definitions of fault signatures by human analysts. The state-of-the-art diagnostics methods have in common that, for each of them, a human analyst has investigated and designed a particular feature (fault signature) to be extracted from the vibration measurements. Each feature has been defined so as to capture a particular aspect that starts to build up in the time- or frequency domain signals when an incipient fault starts to evolve and intensify in an originally healthy component. For instance, local surface damage on a gear tooth is typically diagnosed based on changes in the residual signal obtained after the gear mesh frequencies and harmonics have been removed. These feature-engineering approaches have multiple disadvantages as discussed above. They require time-intensive handcrafting of fault diagnostics features and detailed knowledge of the monitored component. Therefore, they lack scalability with the increasing number of monitored turbines, with different component types and configurations present, each of which with its own characteristic frequencies. Fault signatures defined by human analysts can result in biased and imprecise decision boundaries in the fault diagnostics process.

Wee presented a novel accurate fault diagnostics framework for wind turbine gearboxes that overcomes these disadvantages and can be easily incorporated into condition monitoring software or CMS systems for autonomous fault diagnostics decision support. It is based on high-frequency vibration measurements from multiple accelerometers and monitored components. The proposed two-stage framework combines autonomous data-driven learning of fault signatures and health state classification based on convolutional neural networks and isolation forests. In stage one of the presented approach, an isolation forest algorithm detects anomalous component health states based on the features that have been automatically learnt and extracted from the gearbox component spectrograms. This is particularly suitable for operators and monitoring centers that do not have access to sufficient amounts of accelerometer measurements from gearbox fault events. On the other hand, the availability of such observations is required in stage two which involves a multi-label classification by fault types based on spectrogram features extracted from past fault observations.

We have demonstrated and tested the proposed fault diagnostics framework by application to gearbox vibration measurements from two wind turbine drivetrains. The case study performed to this end used accelerometer measurements from a test rig measurement campaign for three different fault types and achieved high fault diagnostics accuracy.

Unlike the state-of-the-art approaches [8-11], the presented method enables automated feature learning and extraction without a human analyst. As demonstrated, given suitable training data, accurate fault diagnosis is possible without any human feature engineering and without the need for storing thousands of spectral characteristics and threshold values to be predefined by monitoring center staff for every turbine. Moreover, the presented fault diagnostics approach does not require detailed knowledge of the gearbox type, manufacturer, composition, gear dimensions, teeth numbers, characteristic bearing frequencies, and so on. Therefore, it can be applied to arbitrary gearbox types, composition and manufacturers. In summary, the proposed framework is advantageous over the state-of-the-art approaches, like monitoring of spectral lines and other characteristic metrics, in that the fault diagnosis features are learnt by the algorithm, in that no gearbox-type-specific diagnostics expertise and no corresponding human feature engineering are required. Moreover, it is not restricted to predefined frequencies or spectral ranges but monitors the full vibration frequency spectrum of interest.

**Funding:** The support of the present work through a grant from the Swiss innovation agency Innosuisse is gratefully acknowledged.

**Acknowledgments:** The author thanks Shawn Sheng from the National Renewable Energy Laboratory for sharing his gearbox accelerometer measurement data for this study. The author also thanks colleagues at WinJi AG for fruitful discussions.

**Conflicts of Interest:** The author declares no conflict of interest. The funders had no role in the design of the study; in the collection, analyses, or interpretation of data; in the writing of the manuscript, or in the decision to publish the results.